\documentclass[runningheads]{llncs}

\usepackage{eccv}

\usepackage{eccvabbrv}

\usepackage{graphicx}
\usepackage{booktabs}

\usepackage{colortbl}
\usepackage{xcolor}
\usepackage{multirow}
\usepackage{wrapfig}
\usepackage{capt-of}
\usepackage[ruled,vlined]{algorithm2e}

\definecolor{bestcolor}{RGB}{255,180,180}
\definecolor{secondcolor}{RGB}{255,220,160}

\newcommand{\bestnum}[1]{%
  \begingroup
  \setlength{\fboxsep}{2.2pt} 
  \colorbox{bestcolor}{$\displaystyle #1$}%
  \endgroup}

\newcommand{\secondnum}[1]{%
  \begingroup
  \setlength{\fboxsep}{2.2pt}
  \colorbox{secondcolor}{$\displaystyle #1$}%
  \endgroup}

\definecolor{cvprblue}{rgb}{0.21,0.49,0.74}

\usepackage[accsupp]{axessibility}  

\usepackage[hidelinks]{hyperref}

\begin{document}

\title{StreamTalk: Streaming Co-Speech Gesture Generation with Key-Pose Anchoring} 

\titlerunning{StreamTalk}

\author{Xiangyue Zhang\inst{1}\thanks{Equal contribution.} \and
Jianfang Li\inst{2}$^{\star}$ \and
Jiaxu Zhang\inst{3} \and
Kaixing Yang\inst{4} \and
Steven Hoi\inst{2}}

\authorrunning{X.~Zhang et al.}

\institute{The University of Tokyo \and
Alibaba Group \and
Nanyang Technological University \and
Renmin University of China}

\maketitle

\begin{figure}[!ht]
\centering
\vspace{-0.3em}
\includegraphics[width=\textwidth]{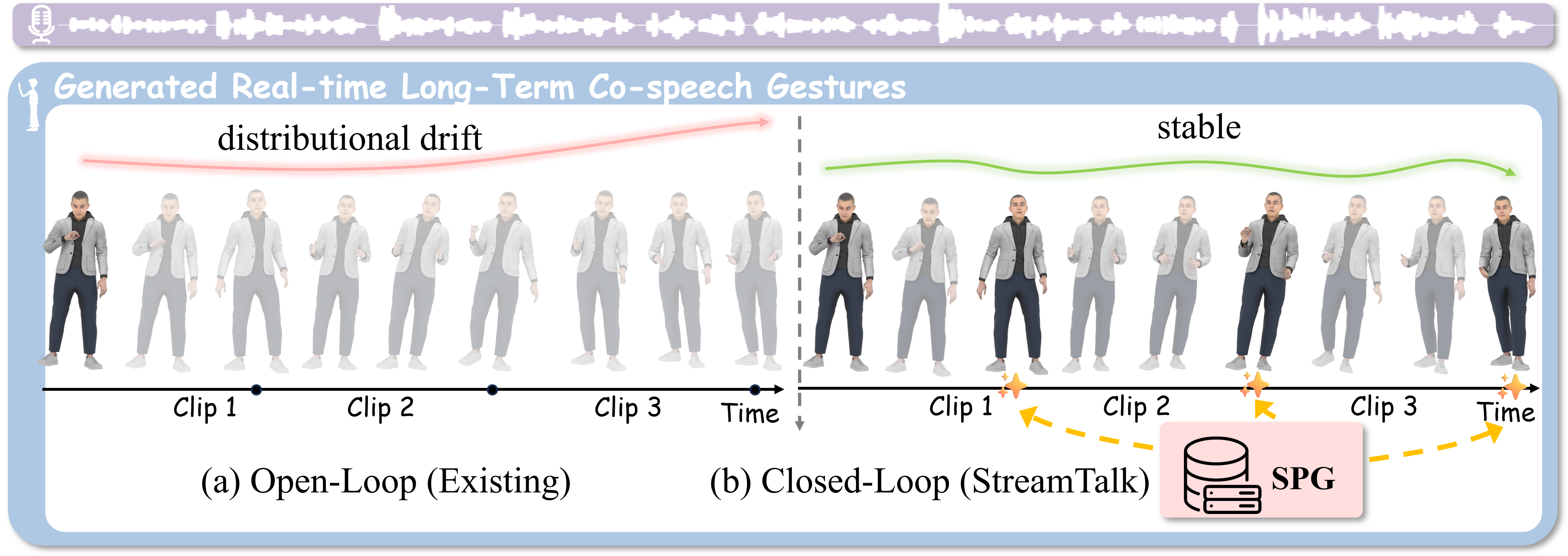}
\caption{
\textbf{(a)} Existing open-loop streaming methods generate each clip conditioned only on past context. Without forward constraints, small errors accumulate across clips, causing \emph{distributional drift}---motion that progressively departs from the natural pose distribution.
\textbf{(b)} StreamTalk introduces closed-loop streaming generation: our Streaming Pose-Guided Generation (SPG) module retrieves a plausible key pose from a motion database at each clip boundary and uses it as a destination anchor, keeping the generated motion stable over long horizons.
}
\label{fig:teaser}
\vspace{-20pt}
\end{figure}

\begin{abstract}
Real-time co-speech gesture generation requires producing 3D motion clip by clip as speech streams in. Existing methods are fundamentally \emph{open-loop}: each clip is synthesized conditioned only on past context, with no mechanism to verify or correct trajectory plausibility. Small per-clip errors therefore compound silently, causing the well-known drift problem---motion that gradually deviates from natural distributions over minute-scale horizons. We make a key observation: drift stems not from poor local motion quality---modern diffusion models already produce convincing short clips---but from the absence of \emph{forward constraints} that tell each clip where it should arrive. Supplying even a single plausible key pose at a clip's tail as a ``destination anchor'' is sufficient to suppress drift dramatically. Building on this insight, we propose StreamTalk, a \emph{closed-loop} streaming framework that introduces a periodic generate--retrieve--refine feedback cycle. At inference, Streaming Pose-Guided Generation (SPG) first produces a coarse clip, retrieves a plausible tail key pose from a speaker-specific motion database, and refines the clip with this anchor before passing it to the next window. To enable the model to exploit such sparse anchors effectively, we introduce Stochastic Anchor Masking (SAM) during training, which independently masks random pose and translation frames so the model learns to inpaint complete motion from partial boundary conditions. A part-aware DiT architecture further disentangles hand, body, and translation streams to prevent global displacement from interfering with local articulation. Extensive experiments on the BEAT2 benchmark demonstrate that StreamTalk achieves state-of-the-art motion quality (FGD), significantly suppresses long-horizon drift compared to open-loop baselines, and runs in real time at 76\,FPS---enabling practical minute-scale streaming co-speech gesture generation. Project page: \url{https://xiangyue-zhang.github.io/StreamTalk/}.
\end{abstract}
    
\section{Introduction}
\label{sec:intro}

Generating natural 3D co-speech gestures is essential for virtual presenters, telepresence avatars, and interactive game characters \cite{zhang2024speech,zhang2024semantic,pan2024expressive,alexanderson2023listen}.
In all of these applications, speech arrives as a continuous stream---users speak in real time, and the corresponding body motion must be produced immediately, clip by clip, rather than after the entire utterance is available.
This \emph{streaming} requirement rules out offline methods that assume access to the full audio and fundamentally shapes the design of the generator.

As illustrated in~\cref{fig:teaser}(a), existing streaming approaches, whether built on VQ-VAE tokenizers~\cite{van2017neural,liu2024emage,zhang2025semtalk,chen2024enabling,tang2026megadance,liu2022disco} or diffusion models~\cite{chen2024diffsheg,yang2023diffusestylegesture,ao2023gesturediffuclip,yang2025flowerdance,chhatre2024emotional,ng2026sarah}, share a common \emph{open-loop} pattern: each clip is synthesized conditioned only on past context (typically the tail frames of the previous clip), and the result is directly forwarded as the seed for the next clip.
The critical weakness of this design is the absence of any verification or correction mechanism: once a clip is produced, small errors in pose or trajectory are locked in and propagated to all subsequent clips.
Over minute-scale horizons, these errors compound into \emph{distributional drift}---the generated motion progressively departs from the natural pose distribution, producing unnatural trajectories and rhythmic desynchronization.
Increasing the overlap between clips or extending the context window can smooth local boundaries, but neither addresses the root cause: an open-loop system has no way to pull a drifting trajectory back toward the plausible distribution.

A closer examination reveals that drift is \emph{not} a local quality problem---modern diffusion models already produce convincing short clips.
The real bottleneck is the lack of a \emph{forward constraint}: each clip knows where it starts (from the previous clip's tail) but has no information about where it should arrive.
Without a plausible destination, the generator is free to wander in any direction that locally satisfies the audio condition.
We find that supplying even a \emph{single} plausible key pose at the clip's tail---a ``destination anchor'' retrieved from a speaker-specific motion database---is sufficient to suppress drift dramatically.
The resulting motion maintains stable quality over long horizons, whereas open-loop variants show progressive degradation.
Conversely, inserting too many anchors per clip \emph{hurts} quality, confirming that drift is a direction problem rather than a per-frame error accumulation: the model needs one waypoint to aim for, not a dense correction grid.

Building on this insight, we propose StreamTalk, a \emph{closed-loop} streaming framework for co-speech gesture generation (\cref{fig:teaser}(b)).
At inference, our Streaming Pose-Guided Generation (SPG) module implements a \emph{generate--retrieve--refine} feedback cycle at each clip boundary: a coarse clip is first produced, its tail pose is matched against a speaker-specific motion database to retrieve the nearest key pose as a destination anchor, and the clip is then refined with this anchor before being forwarded to the next window.
This discrete-time feedback---operating at clip boundaries rather than frame-by-frame---strikes a balance between correction frequency and computational efficiency, analogous to periodic recalibration in navigation systems.
To enable the model to exploit such sparse anchors effectively, we introduce Stochastic Anchor Masking (SAM) during training: pose and translation frames are independently masked at random, so the model learns to \emph{inpaint} complete motion from partial boundary conditions---exactly the capability SPG requires at inference.
A part-aware DiT architecture further separates hand, body, and translation streams, preventing global displacement from interfering with local articulation learning.

Our contributions are as follows:
\begin{itemize}
\item We identify \emph{distributional drift} as an inherent consequence of open-loop streaming architectures and trace its root cause to the absence of forward constraints. We propose \emph{closed-loop} streaming generation as a principled solution, where retrieval-based feedback periodically anchors the generation trajectory back to the plausible distribution.

\item We instantiate this paradigm in StreamTalk: SPG closes the loop through a generate--retrieve--refine cycle at each clip boundary, while SAM trains the model to inpaint motion from sparse anchors so that it can fully exploit the retrieved key poses. A single destination anchor per clip is sufficient for effective correction.

\item Extensive experiments on the BEAT2 benchmark demonstrate state-of-the-art motion quality, significantly suppressed long-horizon drift compared to open-loop baselines, and real-time inference speed, enabling practical minute-scale streaming co-speech gesture generation.
\end{itemize}

\section{Related Work}

\noindent \textbf{Co-speech Gesture Generation.}
Contemporary co-speech gesture models fall into two families.
\emph{VQ-VAE methods}~\cite{lee2022autoregressive, yi2023generating,liu2024emage,liu2024towards,zhang2025semtalk,zhang2025echomask,chen2024enabling} encode motion into discrete tokens decoded autoregressively, while \emph{diffusion methods}~\cite{yang2023diffusestylegesture,chen2024diffsheg,ao2023gesturediffuclip,cheng2025hologest,zhi2023livelyspeaker,liu2025gesturelsm,zhang2025mitigating,ao2023gesturediffuclip} produce smoother local motion via continuous denoising.
Recent work also studies personalized speakers and human-centric audio-video avatars~\cite{zhang2026personagesture,cheng2026unison,song2026interactiveavatar}, which are complementary to our focus on long-horizon online motion stability.
Crucially, \emph{both families operate in open-loop fashion when streaming}: each clip is generated from past context alone without trajectory verification, so per-clip errors accumulate into long-term drift.
DiffSHEG~\cite{chen2024diffsheg} applies inpainting for boundary smoothing, but its manually designed masks cannot prevent global trajectory divergence.
StreamTalk addresses this by \emph{closing the loop}: a retrieved key pose serves as a forward constraint that corrects the trajectory after each clip.

\noindent \textbf{Long-term Motion Generation.}
Autoregressive rolling~\cite{li2017auto,yang2023synthesizing,shafir2023human} conditions each window on the previous output but is inherently open-loop.
Segment-wise blending~\cite{athanasiou2022teach,wang2022neural,li2024lodge,zhuo2025infinidreamer} fuses overlapping segments offline, which is too costly for real-time streaming.
Recent work scales to longer horizons via memory compression~\cite{zhang2024infinimotion}, score distillation~\cite{zhuo2025infinidreamer}, causal latent diffusion~\cite{xiao2025motionstreamer}, or retrieval augmentation~\cite{mughal2025retrieving}.
While these designs slow drift, they remain open-loop.
StreamTalk instead uses retrieval as \emph{feedback correction}: the retrieved tail pose serves as a destination anchor for clip refinement, closing the loop.

\noindent \textbf{Broader Human-Centric Motion and Multimodal Modeling.}
Human motion research also spans motion style transfer and fine-grained motion-language retrieval~\cite{chen2025astf,chen2026beyond}, controllable text-to-motion generation~\cite{zhang2025towards,wang2025text,wang2026generating}, topology-agnostic character animation and retargeting~\cite{zhang2026semantic,zhang2024tapmo,zhang2023skinned}, and music-driven dance generation or retrieval~\cite{tang2026megadance,yang2025flowerdance,yang2024beatdance,yang2024codancers,yang2025cohedancers,yang2026mace,yang2026omnidance}.
Other lines study skeleton-based action understanding and multimodal benchmarks beyond human gesture synthesis~\cite{zhang2025skeletonx,zhang2025skeletonmix,zhang2025robust,wang2023neural,cheng2025owlsight,xiao2026geommb}.
These works broaden the modeling tools for structured motion and multimodal perception, but most target offline generation, retrieval, recognition, or non-speech interaction.
StreamTalk instead targets chunk-level online co-speech gesture generation, where each generated window must remain plausible before future speech or future motion is available.

\section{Method}

\subsection{Overview}

We formulate streaming co-speech gesture generation as \emph{sequential motion inpainting}.
At each time window, the model receives two boundary conditions---head frames inherited from the previous clip and a tail key pose retrieved from a plausible motion database---and inpaints the frames in between.
Chaining these inpainting steps produces an arbitrarily long motion sequence while the retrieved tail anchor periodically corrects the generation trajectory, closing the loop that conventional streaming methods leave open.

\noindent \textbf{Speech and Motion Representation.}
The input speech is encoded by a WavLM encoder~\cite{chen2022wavlm} into frame-level embeddings $\mathbf{w}\!\in\!\mathbb{R}^{L\times 1024}$.
Following DiffSHEG~\cite{chen2024diffsheg}, a 4-layer Transformer regresses facial features $\mathbf{f}$ from $\mathbf{w}$, and both are concatenated to form audio--facial features $\mathbf{a}=\text{Concat}(\mathbf{w},\mathbf{f})$.
The target motion $\mathbf{x}\!\in\!\mathbb{R}^{L\times(55\times6+3)}$ follows the SMPL-X convention~\cite{pavlakos2019expressive} with 6D joint rotations~\cite{zhou2019continuity} and root translation.

\noindent \textbf{Generative backbone.}
We adopt the flow-matching objective~\cite{lipman2022flow}, where the model $f_\theta$ directly predicts the clean motion $\mathbf{x}_1$ from a linearly interpolated sample $\mathbf{x}_t=(1-t)\mathbf{x}_0+t\mathbf{x}_1$, $t\sim\mathcal{U}(0,1)$.
Training and loss details are given in Sec.~\ref{sec:sam} and~\ref{sec:loss}.

\begin{figure}
    \centering
    \includegraphics[width=0.9\textwidth]{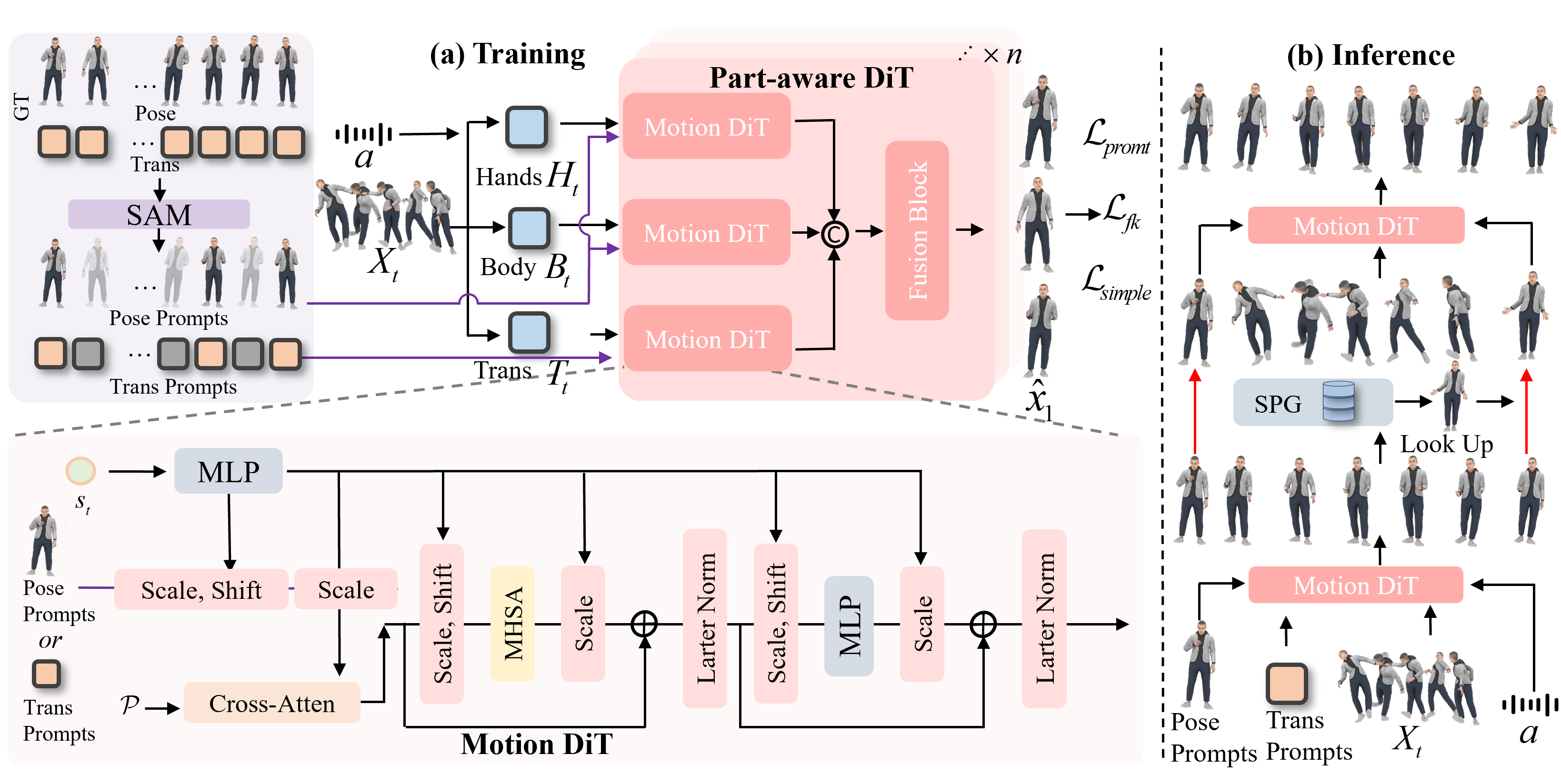}
    \caption{\textbf{Architecture of StreamTalk.}
(a)~\textbf{Training.} SAM independently masks random pose and translation frames from ground-truth motion to form sparse Pose and Trans prompts.
The noised sequence $\mathbf{X}_t$ is split into hands $\mathbf{H}_t$, body $\mathbf{B}_t$, and translation $\mathbf{T}_t$ and processed by a Part-aware DiT with three branches; outputs are fused to predict $\hat{\mathbf{X}}_1$.
This teaches the model to \emph{inpaint} complete motion from partial boundary conditions.
(b)~\textbf{Inference.} SPG forms a closed-loop \emph{generate--retrieve--refine} cycle: each clip is first generated, then its tail is matched against a speaker-specific motion database to retrieve a plausible key pose anchor; the clip is refined with this anchor before its tail frames seed the next window.
}
    \vspace{-5pt}
    \label{fig:method}
\end{figure}
\subsection{Part-aware DiT}
\label{sec:dit}

Rather than processing all joints in a single stream, the Part-aware DiT decomposes motion into three branches---\emph{Hands} ($\mathbf{H}_t$), \emph{Body} ($\mathbf{B}_t$), and \emph{Translation} ($\mathbf{T}_t$)---each implemented as a residual Transformer block with FiLM-style~\cite{perez2018film} scale--shift modulation conditioned on the style vector:
\begin{equation}
s_t=\text{Concat}\!\big(\text{MLP}(\text{PID}),\;\text{MLP}(t)\big).
\end{equation}

A cross-attention layer links each branch-specific feature $\mathcal{P}\in\{H_t,B_t,T_t\}$ with the corresponding prompt (pose or translation), and multi-head self-attention captures temporal dependencies within each part.
The three branch outputs are concatenated and fused through a lightweight attention-based fusion block that enables controlled cross-part coordination.
This disentangled design serves two purposes.
First, it prevents global displacement from corrupting local articulation learning (see ablation in Sec.~\ref{sec:ablation}).
Second, it is a prerequisite for SPG (Sec.~\ref{sec:spg}): because SPG retrieves only \emph{pose} anchors from the database (translation is context-dependent and cannot be borrowed across clips), the model must be able to accept pose and translation prompts through separate channels.

\subsection{Streaming Pose-Guided Generation (SPG)}
\label{sec:spg}

SPG is the core inference mechanism that transforms streaming generation from an open-loop to a closed-loop process.
As illustrated in Figure~\ref{fig:spg} and Algorithm~\ref{alg:spg}, each 60-frame clip is produced through a \emph{generate--retrieve--refine} cycle.

\begin{center}
    \centering
    \includegraphics[width=0.52\textwidth]{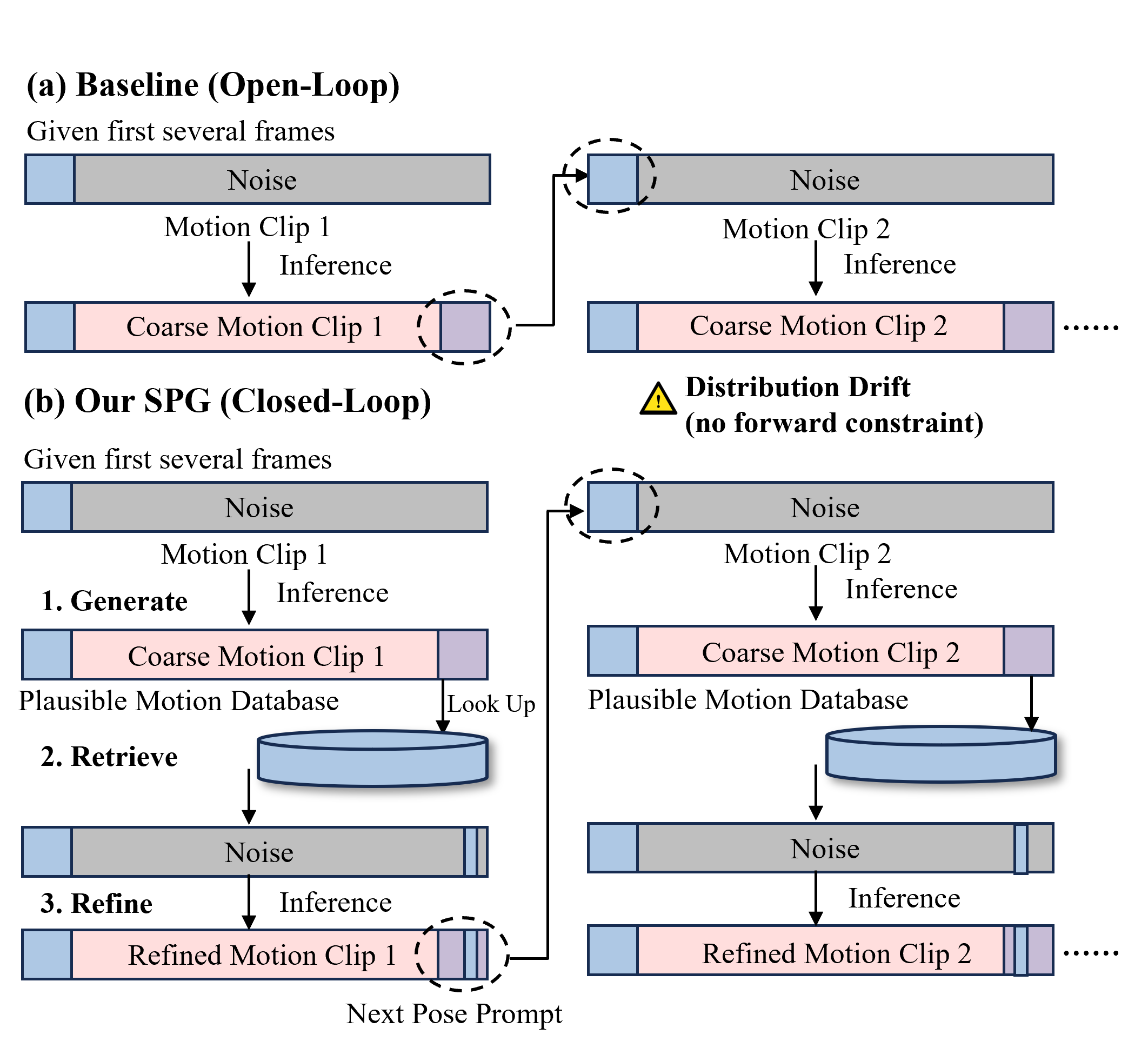}
    \captionof{figure}{\textbf{Open-loop vs.\ closed-loop inference.}
(a)~Open-loop errors accumulate into drift.
(b)~SPG closes the loop with a retrieved destination anchor.}
    \vspace{-8pt}
    \label{fig:spg}
\end{center}

\begin{algorithm}[H]
\caption{Closed-Loop Inference with SPG}
\label{alg:spg}
\footnotesize
\DontPrintSemicolon
\KwIn{Audio $\mathbf{a}$, pose prompt $\mathbf{P}_{\!prompt}$, step size $N$, condition $r$}
\KwOut{Generated motion $\mathbf{x}_1$}
$\mathbf{x}_0 \sim \mathcal{N}(0, I)$;\enspace $\mathbf{x}_t \leftarrow \mathbf{x}_0$;\enspace $h \leftarrow 1/N$;\enspace $\mathrm{stages} \leftarrow [\mathbf{x}_0]$ \\[3pt]
\textbf{1.\ Generate} (forward integration):\\
\For{$i = 0$ \KwTo $N$}{
    $\mathbf{x}_1 \leftarrow f_\theta(\mathbf{x}_t,\, i\!\cdot\! h,\, P_{prompt},\, \mathbf{a},\, r)$\\
    $\mathbf{x}_t \leftarrow \mathbf{x}_0\,(1 \!-\! (i{+}1)h) + \mathbf{x}_1\,(i{+}1)h$;\enspace append $\mathbf{x}_t$ to stages\\
}
\textbf{2.\ Retrieve} (key-pose feedback): $P_{prompt} \leftarrow \mathrm{Retrieve}(P_{prompt},\, \mathbf{x}_1,\, \text{database})$ \\[3pt]
\textbf{3.\ Refine} (corrective integration): $\mathbf{x}_t \leftarrow \mathrm{stages}[\lfloor N/2 \rfloor]$\\
\For{$i = \lfloor N/2 \rfloor$ \KwTo $N$}{
    $\mathbf{x}_1 \leftarrow f_\theta(\mathbf{x}_t,\, i\!\cdot\! h,\, P_{prompt},\, \mathbf{a},\, r)$\\
    $\mathbf{x}_t \leftarrow \mathbf{x}_0\,(1 \!-\! (i{+}1)h) + \mathbf{x}_1\,(i{+}1)h$\\
}
\Return{$\mathbf{x}_1$}
\end{algorithm}


\noindent \textbf{Generate.}
Starting from Gaussian noise $\mathbf{x}_0\!\sim\!\mathcal{N}(0,I)$, the Part-aware DiT integrates the flow-matching ODE over $N$ steps to produce a coarse clip $\hat{\mathbf{x}}_1$, conditioned on the audio--facial feature $\mathbf{a}$, the style vector $s_t$, and the current pose prompt (the tail frames of the previous clip).
All intermediate states $\{\mathbf{x}_{ih}\}$ are cached for the refinement stage.

\noindent \textbf{Retrieve.}
The coarse clip is then verified against a \emph{plausible motion database}.
This database is constructed offline from the training set: for each speaker, we extract all per-frame pose vectors (joint rotations only, excluding root translation) and store them indexed by speaker ID.
Translation is deliberately excluded because global trajectory is context-dependent---a pose observed during walking cannot be meaningfully transplanted to a stationary scene---whereas local articulation patterns (hand shapes, arm configurations) are largely trajectory-invariant and can be safely borrowed across clips.
The database is lightweight and constructed efficiently from training data.
Per-speaker databases are maintained independently to preserve individual motion styles.
When speaker identity is unavailable at inference, a global database merging all speakers is used as a fallback.

Before matching, both the generated tail and all database candidates are projected into \emph{joint space} via forward kinematics (FK).
Comparing in joint space rather than rotation space is critical: small angular errors in proximal joints (e.g., shoulder) cascade into large positional errors at distal joints (e.g., fingertips) along the kinematic chain, so rotation-space distances underestimate perceptually salient discrepan\-cies.
Joint-space matching captures the actual spatial deviation perceived by viewers.
We select the nearest candidate whose joint-space distance falls below a threshold $\theta_p$ and use it to replace the tail key-pose frame in $\hat{\mathbf{x}}_1$, restoring the original translation and global rotation:
\begin{equation}
P_{prompt} \leftarrow \text{Retrieve}(P_{prompt},\, \hat{\mathbf{x}}_1,\, \text{database}).
\end{equation}
A single tail key pose is sufficient: drift is a \emph{direction} problem---the model needs one waypoint to aim for, not a dense correction grid.
We validate this design choice in Sec.~\ref{pos_num}.

\noindent \textbf{Refine.}
With the updated prompt, the model re-integrates from the cached midpoint $\mathbf{x}_{N/2}$ rather than from scratch, halving the cost of the refinement pass.
The refined clip's tail frames are then forwarded as the pose prompt for the next window, completing the feedback loop.

\subsection{Stochastic Anchor Masking (SAM)}
\label{sec:sam}

SPG requires the model to generate complete motion conditioned on sparse boundary anchors---a handful of visible head frames plus a single retrieved tail key pose.
Standard training, where the model always sees the full ground-truth prompt, does not prepare it for this partially observed input.
SAM bridges this gap by introducing the same sparse-anchor pattern during training, so the model learns to \emph{inpaint} motion from partial boundary conditions---exactly the capability SPG exploits at inference.
In essence, SAM is a \emph{training-inference alignment mechanism}: it ensures that the sparse-anchor pattern the model sees during training matches what SPG provides at inference, preventing train-test mismatch.

Concretely, given ground-truth motion $\mathbf{X}_1$, SAM decomposes each frame into joint rotations $\mathbf{P}$ (pose) and root translation $\mathbf{T}$, then applies two \emph{independent} stochastic binary masks $\mathbf{M}_p$ and $\mathbf{M}_t$ to form sparse pose prompts $P_{\mathit{prompt}}\!=\!\mathbf{M}_p\!\odot\!\mathbf{P}$ and translation prompts $T_{\mathit{prompt}}\!=\!\mathbf{M}_t\!\odot\!\mathbf{T}$.
As shown in Figure~\ref{fig:method}(a), pose prompts feed into the Body and Hands branches, while translation prompts feed into the Translation branch.

The masks are independent because pose and translation follow different temporal rhythms in natural motion---a speaker may gesticulate while standing still, or walk without moving the hands.
Independent masking lets the model learn these two rhythms separately.
It also mirrors the SPG design: at inference, SPG retrieves only pose anchors (not translation), so the model must handle the case where pose frames are partially revealed but translation frames are not.

\subsection{Training Objective}
\label{sec:loss}

The model $f_\theta$ is trained to predict the clean motion $\mathbf{X}_1$ from the interpolated sample $\mathbf{X}_t$ under audio--facial condition $\mathbf{a}$, style vector $s_t$, and the SAM-generated prompts.
The primary loss is the flow-matching regression:
\begin{equation}
\mathcal{L}_{\text{simple}} =
\mathbb{E}\big[\|f_\theta(\mathbf{X}_t,P_{prompt},\mathbf{a},s_t,T_{prompt}) - \mathbf{X}_1\|_2^2\big].
\end{equation}

To preserve anatomical structure, a forward-kinematics \cite{isaacs1987controlling} loss matches the 3D joint positions computed from SMPL-X rotations, following \cite{zhang2023remodiffuse}:
\begin{equation}
\mathcal{L}_{fk} =
\frac{1}{L}\sum_{t}
\big\|
\mathrm{FK}(\mathbf{X}_1) -
\mathrm{FK}(\hat{\mathbf{X}}_1^{(t)})
\big\|_2^2,
\end{equation}
where $\hat{\mathbf{X}}_1^{(t)}$ is the model prediction at flow step $t$.
A prompt consistency term aligns the prediction with the visible anchors:
\begin{equation}
\begin{aligned}
\mathcal{L}_{prompt}
&= \frac{1}{L}\sum_{t}
\Big(
\|\mathbf{M}_p^{(t)}\odot(\hat{\mathbf{P}}^{(t)}-\mathbf{P}_{\text{prompt}}^{(t)})\|_2^2 \\
&\qquad\;\;+
\|\mathbf{M}_t^{(t)}\odot(\hat{\mathbf{T}}^{(t)}-\mathbf{T}_{\text{prompt}}^{(t)})\|_2^2
\Big).
\end{aligned}
\end{equation}
The total objective is:
\begin{equation}
\mathcal{L}
=
\mathcal{L}_{simple}
+ \lambda_{fk}\mathcal{L}_{fk}
+ \lambda_{prompt}\mathcal{L}_{prompt},
\end{equation}
where $\lambda_{fk}$ and $\lambda_{prompt}$ balance the structural and prompt consistency terms.

\section{Experiments}

\subsection{Experimental Setup}

\noindent\textbf{Datasets.}
We train and evaluate on the BEAT2 dataset~\cite{liu2024emage}, which provides high-quality SMPL-X-based 3D motion synchronized with audio from 25 speakers.
We follow the official train/validation/test split.
Following~\cite{mughal2025retrieving}, we report results under both the \emph{1-Speaker} setting (speaker ``Scott'') and the \emph{All-Speakers} setting to assess generalization across identities.

\noindent\textbf{Implementation Details.}
Training is conducted on four NVIDIA V100 GPUs (16\,GB) for 1{,}000 epochs with batch size 128, taking approximately 31 hours.
We use the ADAM optimizer with learning rate $1\times10^{-4}$.
Following~\cite{liu2024emage,zhang2025semtalk}, each clip contains 60 frames.
For streaming generation, the first 8 frames initialize the sequence; thereafter the last 8 frames of each clip serve as the pose prompt for the next, with speech segmented using an 8-frame overlap.
Loss weights are set to $\lambda_{fk}=1$ and $\lambda_{prompt}=0.1$.
The denoising step size is $N\!=\!10$; one key pose ($n\!=\!1$) is selected from the tail frames for SPG.

\noindent\textbf{Speaker Identity as Input.}
Following standard practice in co-speech gesture generation~\cite{liu2024emage,zhang2025semtalk,chen2024diffsheg,liu2025gesturelsm}, speaker identity (ID) is provided as input to condition the model on individual motion styles.
All baseline methods also use speaker ID as a style variable.
This is not information leakage but a standard setup: in practical applications (virtual avatars, telepresence), the speaker is known and consistent throughout the session.

\noindent\textbf{Plausible Motion Database Construction.}
For the 1-Speaker setting (Speaker ``Scott''), the database contains 156{,}977 pose frames extracted from the training set, occupying only 16.6\,MB in memory.
For the All-Speakers setting, per-speaker databases are maintained independently to preserve individual motion styles.
Since speaker identity is a standard input (see above), SPG retrieves key poses from the target speaker's database partition at inference---this is a fair comparison, as baseline methods also condition on speaker ID to generate speaker-specific motion.
When speaker identity is unavailable at inference (e.g., unseen speakers in zero-shot settings), SPG falls back to a global database merging all training speakers.

\noindent\textbf{Metrics.}
We evaluate generated motion with five metrics.
\emph{Fr\'echet Gesture Distance} (FGD)~\cite{yoon2020speech} measures distributional similarity between generated and real motions.
\emph{Diversity}~\cite{li2021audio2gestures} is the average L1 distance among different generated samples.
\emph{Beat Consistency} (BC)~\cite{li2021ai} measures temporal alignment between motion and speech beats.
\emph{Velocity distance} (VEL) and \emph{Acceleration distance} (ACC) are the L2 distances between the velocity and acceleration magnitude histograms of generated and ground-truth motions, capturing dynamic fidelity.

\begin{table}[t]
   \centering
   \setlength{\tabcolsep}{3pt}
   \scriptsize
   \begin{minipage}[t]{0.48\textwidth}
   \centering
   \textbf{1 Speaker}\\[2pt]
    \begin{tabular}{lccc}
        \toprule
        \textbf{Method} & \textbf{FGD$\downarrow$} & \textbf{BC$\rightarrow$} & \textbf{DIV$\rightarrow$} \\
        \midrule
        GT & -- & 0.703 & 11.97  \\
        \cmidrule(lr){1-4}
        CaMN~\cite{liu2022beat} & 0.664 & 0.677 & 10.86  \\
        ReMoDiffuse \cite{zhang2023remodiffuse} & 0.702 & 0.824 & 12.46 \\
        DSG~\cite{yang2023diffusestylegesture} & 0.881 & 0.724 & 11.49  \\
        Audio2Photoreal \cite{ng2024audio2photoreal} & 1.02 & 0.550 & \secondnum{12.47}  \\
        LivelySpeaker~\cite{zhi2023livelyspeaker} & 1.180 & 0.666 & 11.28  \\
        Habibie \textit{et al.}~\cite{habibie2021learning} & 0.904 & 0.772 & 8.213  \\
        TalkSHOW~\cite{yi2023generating} & 0.621 & 0.695 & 13.47  \\
        DiffSHEG~\cite{chen2024diffsheg} & 0.899 & \secondnum{0.714} & 11.91 \\
        AMUSE \cite{chhatre2024emotional} & 1.211 & 0.832 & 14.93 \\
        EMAGE~\cite{liu2024emage} & 0.551 & 0.772 & 13.06  \\
        MambaTalk \cite{xu2024mambatalk} & 0.537 & 0.781 & 13.05 \\
        SynTalker \cite{chen2024enabling} & 0.641 & 0.736 & 12.72 \\
        HoloGest \cite{cheng2025hologest} & 0.534 & 0.795 & 14.15  \\
        RAG-GESTURE \cite{mughal2025retrieving} & 0.808 & 0.734 & \bestnum{11.97} \\
        PyraMotion \cite{yinpyramotion} & 0.461 & 0.742 &  13.24 \\
        SemGes \cite{liu2025semges} & 0.447 & - &  - \\
        EchoMask \cite{zhang2025echomask} & 0.462 & 0.774 &  13.37 \\
        GestureLSM \cite{liu2025gesturelsm} & \secondnum{0.409} & \secondnum{0.714} &  13.42 \\
        SemTalk \cite{zhang2025semtalk} & 0.428 & 0.777 & 12.91 \\
        \cmidrule(lr){1-4}
        \textbf{StreamTalk (Ours)} & \bestnum{0.383} & \bestnum{0.704} & 13.18  \\
        \bottomrule
    \end{tabular}
   \end{minipage}
   \hfill
   \begin{minipage}[t]{0.48\textwidth}
   \centering
   \textbf{All Speakers}\\[2pt]
    \begin{tabular}{lccc}
        \toprule
        \textbf{Method} & \textbf{FGD$\downarrow$} & \textbf{BC$\rightarrow$} & \textbf{DIV$\rightarrow$} \\
        \midrule
        GT & -- & 0.477 & 7.29   \\
        \cmidrule(lr){1-4}
        CaMN~\cite{liu2022beat} & 0.512 & 0.200 & 5.58 \\
        ReMoDiffuse \cite{zhang2023remodiffuse} & 1.120 & 0.218 & 5.06 \\
        DSG~\cite{yang2023diffusestylegesture} & 1.174 & 0.734 & 11.12 \\
        Audio2Photoreal \cite{ng2024audio2photoreal} & 0.849 & 0.326 & \secondnum{6.24}  \\
        EMAGE~\cite{liu2024emage} & 0.692 & 0.284 & 6.06  \\
        HoloGest \cite{cheng2025hologest} & 0.646 & 0.803 & 13.53  \\
        EchoMask \cite{zhang2025echomask} & 0.566 & \bestnum{0.495} & 9.30 \\
        RAG-GESTURE \cite{mughal2025retrieving}  & 0.487 & \secondnum{0.514} & 9.94 \\
        SemTalk \cite{zhang2025semtalk} & \secondnum{0.356} & 0.841 & 8.41 \\
        \cmidrule(lr){1-4}
        \textbf{StreamTalk (Ours)} & \bestnum{0.293} & 0.616 & \bestnum{7.27}\\
        \bottomrule
    \end{tabular}
    \vspace{6pt}
    \includegraphics[width=\textwidth]{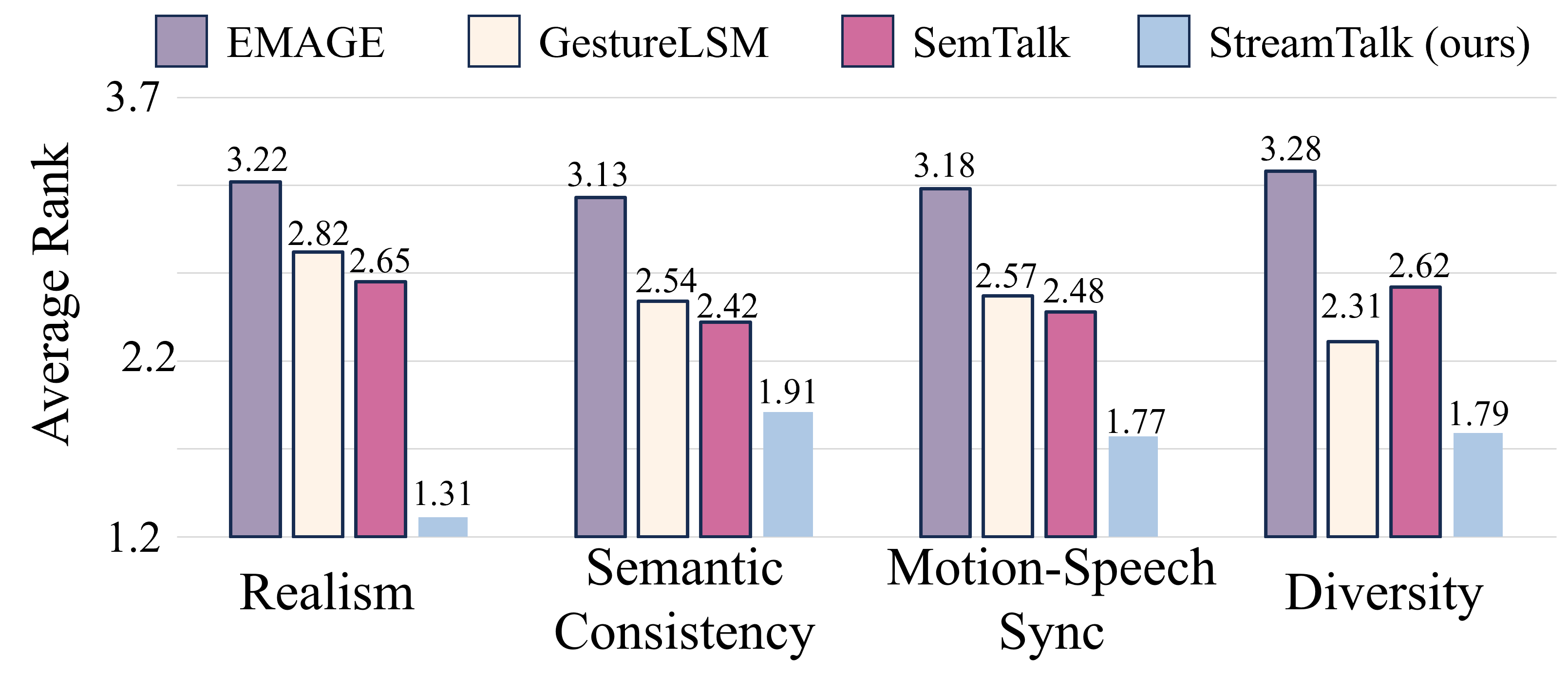}
    \captionof{figure}{\textbf{User study results.} StreamTalk ranks first across all four perceptual criteria.}
    \label{fig:user}
   \end{minipage}
\caption{\textbf{Comparison with state-of-the-art methods on BEAT2.} StreamTalk achieves the best FGD under both settings, with particularly strong generalization across speakers.}
    \vspace{-15pt}
    \label{tab:sota}
\end{table}

\subsection{Comparison with State of the Art}

\noindent \textbf{Quantitative Comparison.}
As shown in Table~\ref{tab:sota}, StreamTalk achieves state-of-the-art FGD under both evaluation settings.
In the \emph{1-Speaker} setting, StreamTalk obtains the lowest FGD, surpassing all prior methods including GestureLSM~\cite{liu2025gesturelsm} and SemTalk~\cite{zhang2025semtalk}, while attaining the closest BC to the ground truth.
Notably, diffusion-based methods (DiffSHEG, DSG, GestureLSM) and VQ-VAE methods (EMAGE, SemTalk, SynTalker) both suffer from open-loop drift in streaming settings; StreamTalk's closed-loop correction benefits both paradigms, as it operates on the output trajectory rather than any specific backbone.

In the more challenging \emph{All-Speakers} setting, StreamTalk outperforms all baselines on FGD, while the Diversity score nearly matches the ground truth.
This indicates that SPG's retrieval-based anchoring does not suppress motion variety---the database provides a plausible \emph{direction} rather than a rigid template, leaving the flow-matching backbone free to generate diverse motions within the corrected trajectory.
These results confirm that the closed-loop design generalizes well across speaker identities.

\begin{figure}[t]
    \centering
    \includegraphics[width=0.9\textwidth]{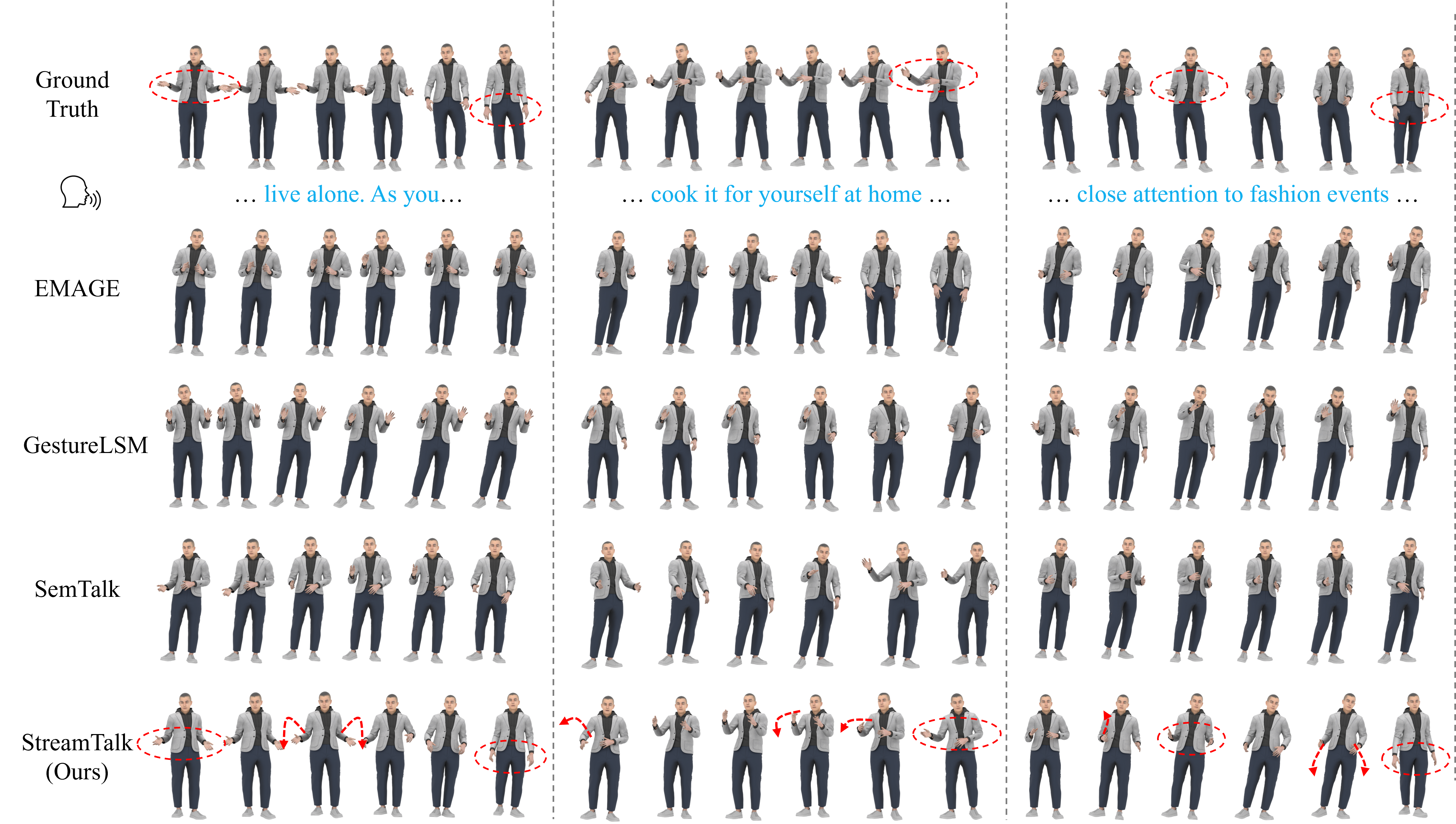}
    \caption{\textbf{Qualitative comparison on BEAT2~\cite{liu2024emage}.} Each column shows gestures aligned with the same speech segment. StreamTalk (bottom) produces smoother, more expressive, and rhythm-aligned gestures with consistent arm spacing, while baselines exhibit motion freezing, desynchronized gestures, or loss of spatial consistency. Red dashed ellipses highlight regions where StreamTalk better preserves rhythmic articulation and continuity.}
    \vspace{-8pt}
    \label{fig:compare}
\end{figure}

\noindent \textbf{Qualitative Comparison.}
We highly recommend readers watch the supplementary demo video.
As shown in Figure~\ref{fig:compare}, StreamTalk produces fluid, temporally coherent gestures that follow prosodic rhythm and maintain balanced arm spacing.
EMAGE~\cite{liu2024emage} tends to produce repetitive, low-amplitude arm swings that lose expressiveness after the first few seconds, a symptom of VQ-VAE codebook collapse under streaming conditions.
GestureLSM~\cite{liu2025gesturelsm} generates locally smooth clips but exhibits visible spatial drift at clip boundaries---the arms gradually shift upward or sideways, breaking the natural resting pose envelope.
SemTalk~\cite{zhang2025semtalk} preserves semantic coherence for short segments but suffers from pose freezing during pauses or low-energy speech, where the model defaults to a near-static mean pose rather than producing subtle idle motion.
In contrast, StreamTalk maintains consistent motion amplitude, smooth inter-clip transitions, and natural idle behavior throughout the entire sequence, owing to the periodic trajectory correction provided by SPG.

\noindent \textbf{User Study.}
Twenty-seven participants evaluated 15 shuffled video segments (${\sim}$60\,s each), yielding over 400 rated samples.
For each segment, outputs from EMAGE, GestureLSM, SemTalk, and StreamTalk were presented in random order and ranked on realism, rhythm consistency, motion--speech synchrony, and diversity.
As shown in Figure~\ref{fig:user}, StreamTalk receives the highest overall preference, with statistically significant gains in realism and rhythm consistency under the Wilcoxon signed-rank test ($p < 10^{-3}$).

\subsection{Ablation Studies}
\label{sec:ablation}

\noindent \textbf{Does Closing the Loop Help?}
Table~\ref{tab:ablation_combined}(a) isolates the contributions of SAM and SPG under both evaluation settings.
On the \emph{1-Speaker} setting, adding SAM alone slightly increases FGD, which is expected: SAM is a \emph{training-inference alignment mechanism} whose value is realized only when paired with SPG's sparse anchors at inference.
Adding SPG alone already improves FGD via trajectory correction; combining both yields the best result with a substantial reduction from the baseline.
Under the \emph{All-Speakers} setting, the same complementary pattern holds: combining SAM and SPG achieves the best FGD, confirming that the closed-loop design generalizes across speaker identities.

\begin{table}[!t]
\centering
\scriptsize
\setlength{\tabcolsep}{3.5pt}
\begin{minipage}[t]{\textwidth}
    \centering
    \begin{tabular}{l ccc ccc}
    \toprule
    & \multicolumn{3}{c}{\textbf{1-Speaker (Speaker-2)}} & \multicolumn{3}{c}{\textbf{All Speakers}} \\
    \cmidrule(lr){2-4} \cmidrule(lr){5-7}
    \textbf{Method} & \textbf{FGD$\downarrow$} & \textbf{BC$\rightarrow$} & \textbf{Div$\rightarrow$} & \textbf{FGD$\downarrow$} & \textbf{BC$\rightarrow$} & \textbf{Div$\rightarrow$}\\
    \midrule
    GT & -- & 0.703 & 11.97 & -- & 0.477 & 7.29\\
    \cmidrule(lr){1-7}
    StreamTalk (base) & 0.478 & 0.716 & 12.30 & 0.391 & 0.621 & 7.21 \\
    + SAM & 0.503 & 0.747 & 13.72 & 0.379 & 0.613 & 7.31 \\
    + SPG & 0.455 & 0.695 & 14.24 & 0.353 & 0.607 & 7.49 \\
    + SAM \& SPG & \textbf{0.383} & \textbf{0.704} & 13.18 & \textbf{0.293} & 0.616 & 7.27 \\
    \bottomrule
    \end{tabular}\\[2pt]
    {\scriptsize (a) Component ablation across both evaluation settings.}
\end{minipage}

\vspace{8pt}

\begin{minipage}[t]{0.45\textwidth}
    \vspace{0pt}
    \centering
    \begin{tabular}{lcr}
    \toprule
    \textbf{Method} & \textbf{VEL$\downarrow$} & \textbf{ACC$\downarrow$} \\
    \midrule
    GT            & 0.0     & 0.0 \\
    \cmidrule(lr){1-3}
    EMAGE         & 2.40e2  & 4.03e2 \\
    GestureLSM    & 2.33e2  & 4.30e2 \\
    SemTalk       & 2.28e2  & 3.58e2 \\
    \textbf{Ours} & \textbf{2.15e2}  & \textbf{2.33e2} \\
    \bottomrule
    \end{tabular}\\[2pt]
    {\scriptsize (b) VEL \& ACC distances.}
\end{minipage}
\hfill
\begin{minipage}[t]{0.50\textwidth}
    \vspace{0pt}
    \centering
    \begin{tabular}{lccc}
    \toprule
    \textbf{Variant} & \textbf{FGD$\downarrow$} & \textbf{BC$\rightarrow$} & \textbf{Div$\rightarrow$} \\
    \midrule
    GT      & --     & 0.703 & 11.97 \\
    \cmidrule(lr){1-4}
    Random anchor            & 0.673  & 0.743 & 13.12  \\
    Retrieved anchor         & 0.503  & 0.747 & 12.57  \\
    Retrieved + Linear ref.  & 0.471  & 0.753 & 12.31  \\
    Retrieved + Our ref.     & \textbf{0.383}  & \textbf{0.704} & 13.11 \\
    \bottomrule
    \end{tabular}\\[2pt]
    {\scriptsize (c) Anchor source and refinement strategy.}
\end{minipage}
\caption{\textbf{Ablation and analysis.} (a)~Component ablation on 1-Speaker and All Speakers. (b)~VEL/ACC distances. (c)~Anchor source and refinement strategy.}
\vspace{-12pt}
\label{tab:ablation_combined}
\end{table}

\begin{table}[t]
    \centering
    \footnotesize
    \setlength{\tabcolsep}{4pt}
    \begin{minipage}[t]{0.48\textwidth}
        \vspace{0pt}
        \centering
        \resizebox{\textwidth}{!}{%
        \begin{tabular}{llccc}
            \toprule
             \textbf{Position} & \textbf{Number} & \textbf{FGD$\downarrow$} & \textbf{BC$\rightarrow$} & \textbf{Div$\rightarrow$}\\
            \midrule
             -- & -- & -- & 0.703 & 11.97\\
            \cmidrule(lr){1-5}
             random & 8 & 0.601 & 0.759 & 13.51 \\
             random & 4 & 0.540 & 0.716 & 13.43 \\
             random & 1 & 0.426 & 0.715 & 13.05 \\
             \cmidrule(lr){1-5}
             middle & 8 & 0.643 & 0.720 & 13.24 \\
             middle & 4 & 0.574 & 0.664 & 13.03 \\
             middle & 1 & 0.408 & 0.693 & 13.55 \\
             \cmidrule(lr){1-5}
             tail & 8 & 0.575 & 0.773 & 13.31 \\
             tail & 4 & 0.443 & 0.700 & 13.15 \\
             tail & 1 & \textbf{0.383} & \textbf{0.704} & 13.18 \\
            \bottomrule
        \end{tabular}}\\[2pt]
        {\scriptsize (a) Key pose number and position ablation.}
        \label{tab:position_num}
        \vspace{8pt}
    \end{minipage}
    \hfill
    \begin{minipage}[t]{0.48\textwidth}
        \vspace{0pt}
        \centering
        \resizebox{\textwidth}{!}{%
        \begin{tabular}{lcccc}
            \toprule
            \textbf{Backbone} & \textbf{FGD$\downarrow$} & \textbf{BC$\rightarrow$} & \textbf{Div$\rightarrow$} & \textbf{Params} \\
            \midrule
            GT            & --     & 0.703 & 11.97 & -- \\
            \cmidrule(lr){1-5}
            1-branch      & 0.612  & 0.664 & 12.31 & 70.0\,M \\
            2-branch      & 0.562  & \textbf{0.692} & 13.68 & 71.0\,M \\
            3-branch      & \textbf{0.478}  & 0.716  & \textbf{12.30} & 71.2\,M \\
            \bottomrule
        \end{tabular}}\\[2pt]
        {\scriptsize (b) Part-aware DiT architecture ablation.}
        \label{tab:part_anchor}

        \vspace{8pt}
        \resizebox{0.55\textwidth}{!}{%
        \begin{tabular}{lc}
            \toprule
             & \textbf{\# Frames} \\
            \midrule
            w/o SPG & 387 \\
            w/ SPG & \textbf{39} \\
            \bottomrule
        \end{tabular}}\\[2pt]
        {\scriptsize (c) Self-intersection counts.}
        \label{tab:self_inter}
    \end{minipage}
    \caption{\textbf{Key pose and architecture analysis on Speaker-2.} (a)~A single tail key pose yields the best quality. (b)~The 3-branch design outperforms single-branch with minimal parameter overhead. (c)~Self-intersection frame counts.}
    \vspace{-5pt}
    \label{tab:pose_arch}
\end{table}

\begin{center}
    \centering
    \includegraphics[width=0.5\linewidth]{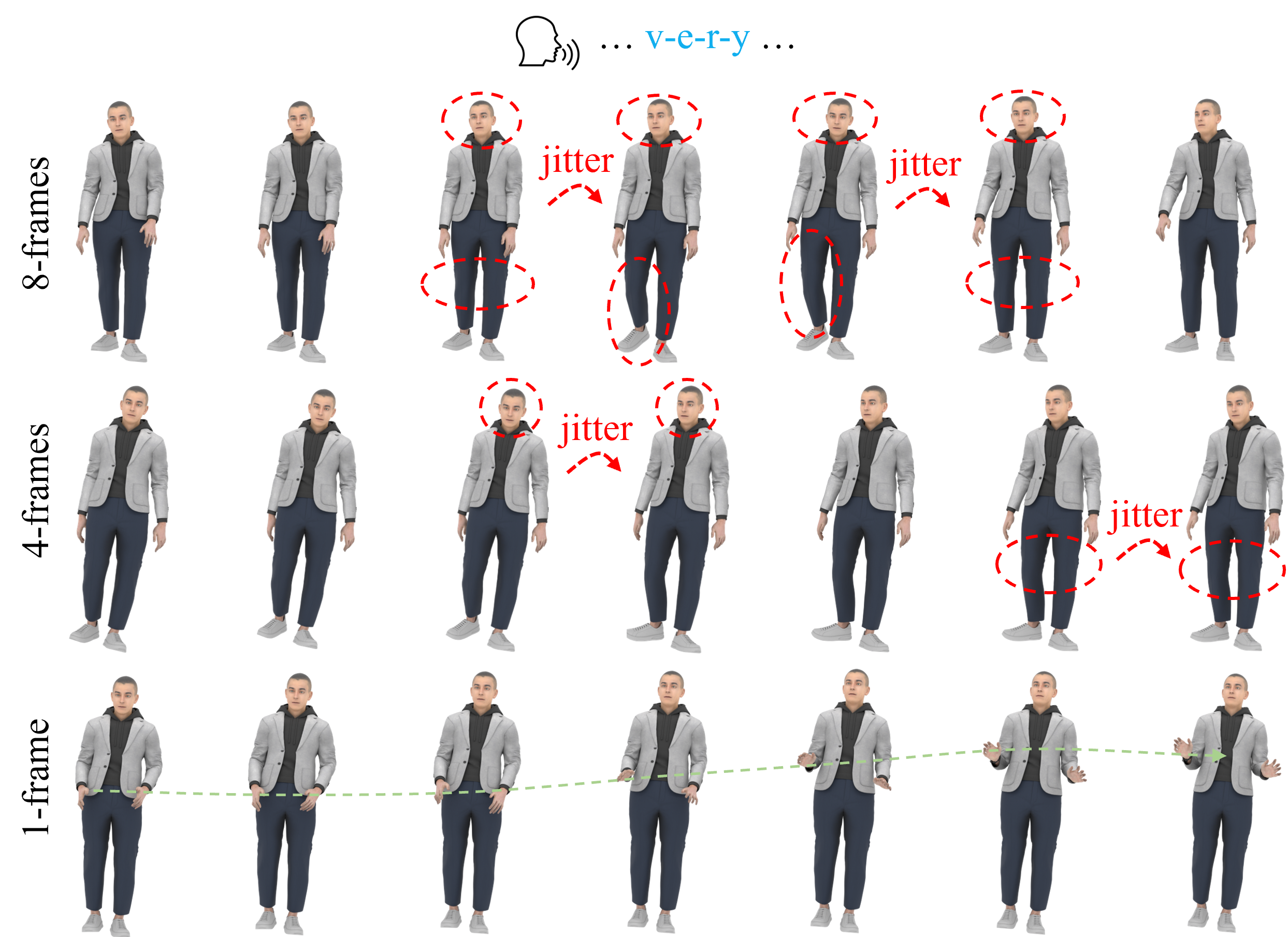}
    \captionof{figure}{\textbf{Effect of key pose count.} Using 8 or 4 anchors introduces visible jitter (red), while a single key pose produces smoother motion.}
    \vspace{-8pt}
    \label{fig:posnum_vis}
\end{center}

\begin{figure}[t]
    \centering
    \begin{minipage}[t]{0.48\textwidth}
        \centering
        \includegraphics[width=\textwidth]{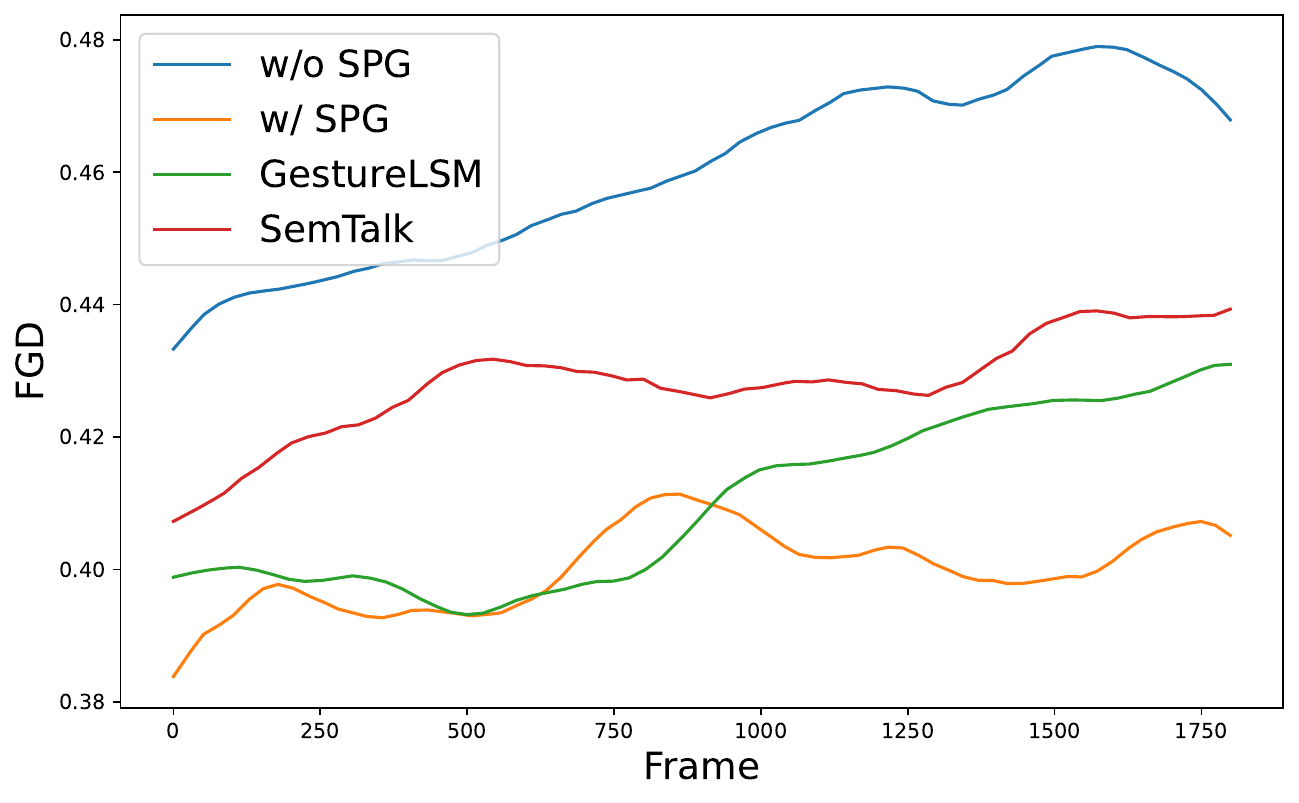}
        \small (a) FGD over time.
    \end{minipage}
    \begin{minipage}[t]{0.48\textwidth}
        \centering
        \includegraphics[width=\textwidth]{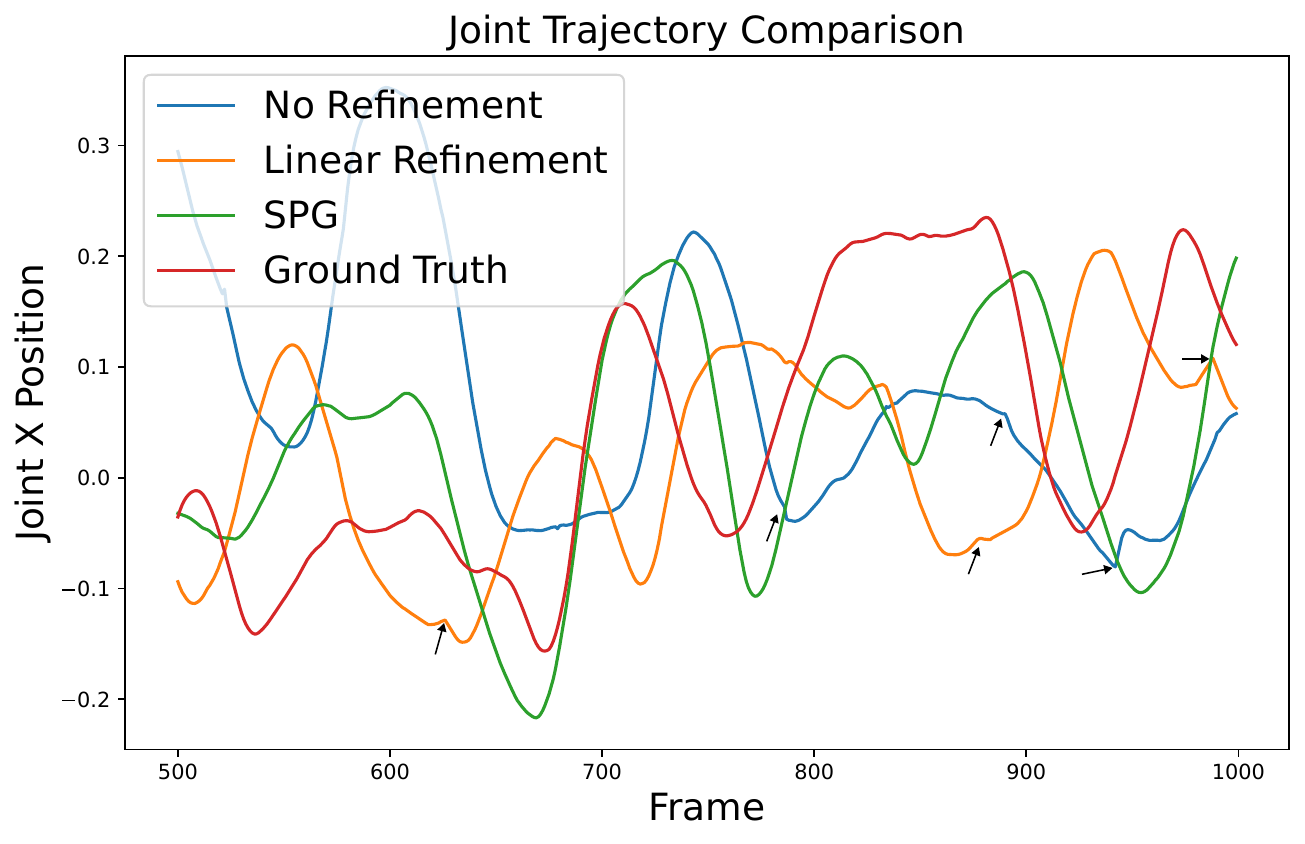}
        \small (b) Joint trajectory comparison.
    \end{minipage}
    \caption{\textbf{Long-horizon stability and refinement analysis.} (a)~Sliding-window FGD across ${\sim}$1{,}800 frames: all baseline methods show upward FGD drift over time, while w/ SPG remains the lowest and most stable. (b)~Joint trajectory of different refinement strategies: No Refinement and Linear Refinement exhibit noticeable jitter, while SPG produces a smooth trajectory closer to the ground truth. Black arrows indicate jittering points.}
    \vspace{-10pt}
    \label{fig:fgd_posnum}
\end{figure}

\begin{figure}[t]
    \centering
    \begin{minipage}[t]{0.6\textwidth}
        \centering
        \includegraphics[width=\textwidth]{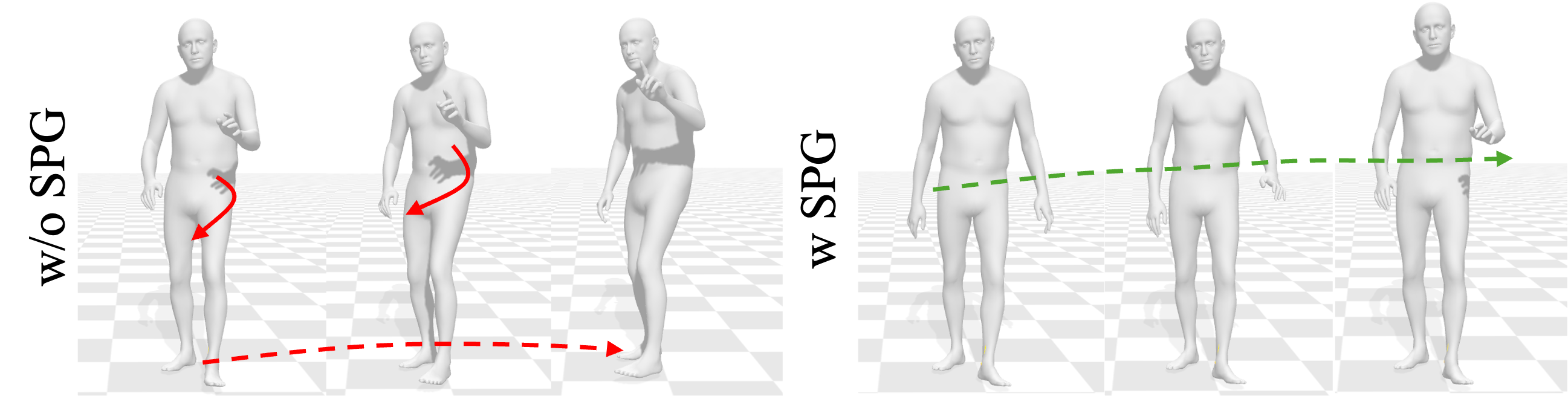}
        \small (a) w/o SPG vs.\ w/ SPG.
    \end{minipage}
    \hfill
    \begin{minipage}[t]{0.38\textwidth}
        \centering
        \includegraphics[width=\textwidth]{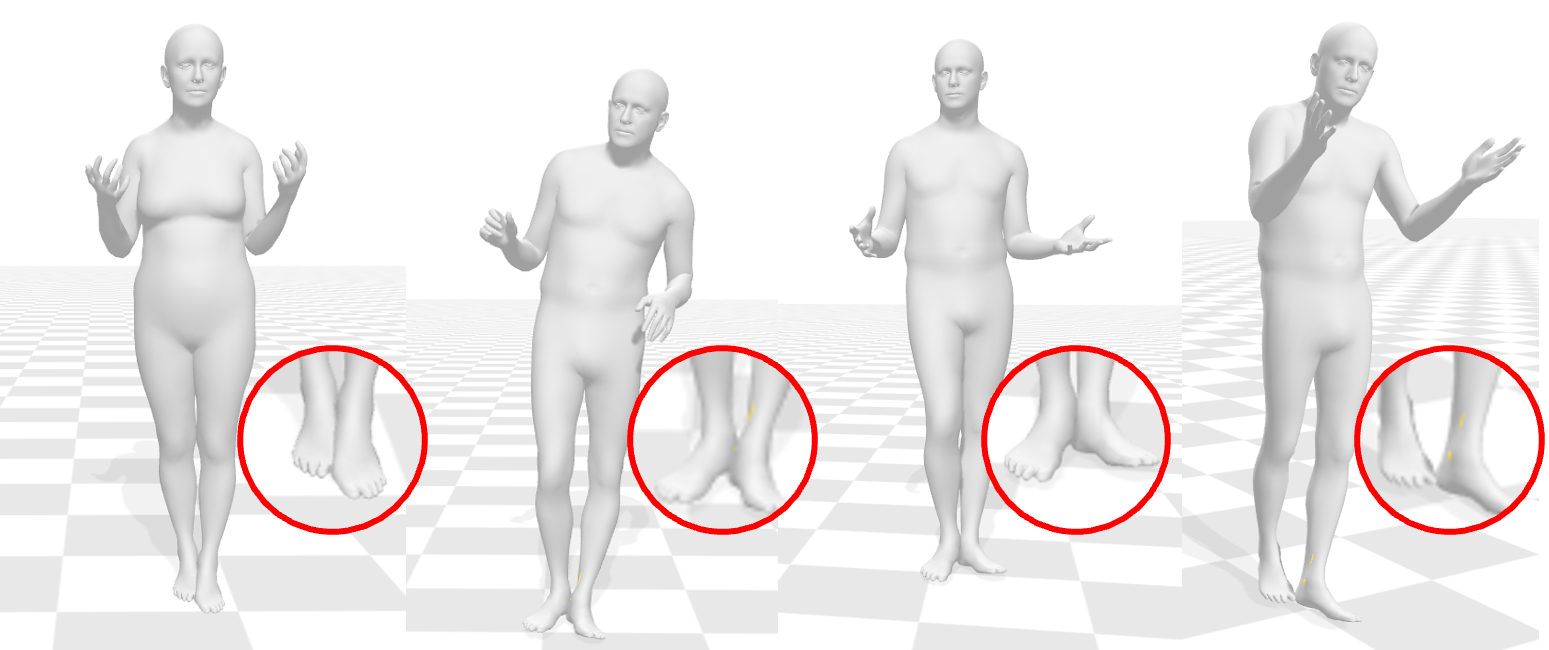}
        \small (b) Self-intersection.
    \end{minipage}
    \caption{\textbf{SPG visual analysis.} (a)~Without SPG, motion shows spatial drift and abrupt corrections; with SPG, trajectories remain smooth. (b)~Self-intersection artifacts are dramatically reduced by SPG.}
    \vspace{-10pt}
    \label{fig:spg_vis}
\end{figure}

\noindent \textbf{Long-Horizon Stability.}
The closed-loop advantage becomes more pronounced over longer sequences.
Figure~\ref{fig:fgd_posnum}(a) plots sliding-window FGD across ${\sim}$1{,}800 frames (${\sim}$60\,s).
All baseline methods exhibit a clear upward FGD trend over time, indicating that open-loop generation---regardless of backbone architecture---accumulates distributional drift as the sequence grows.
In contrast, w/ SPG remains the lowest and most stable throughout the entire sequence, confirming that the periodic key-pose feedback effectively suppresses drift accumulation.
Figure~\ref{fig:spg_vis} provides a visual comparison: without SPG, the generated motion shows visible spatial drift and abrupt corrections, while SPG maintains smooth, consistent trajectories.
Table~\ref{tab:ablation_combined}(b) further shows that StreamTalk achieves the smallest velocity and acceleration histogram distances to GT among all compared methods, indicating more faithful motion dynamics.

Beyond distributional metrics, we examine physical plausibility via self-intersection analysis.
As shown in Figure~\ref{fig:spg_vis} and Table~\ref{tab:pose_arch}(c), open-loop generation produces a significant number of frames with self-intersection artifacts (e.g., arm interpenetration).
SPG dramatically reduces this by an order of magnitude, because the retrieved key poses are drawn from anatomically valid training data, effectively guiding the generation toward physically plausible configurations.

\noindent \textbf{Key Pose Design.} \label{pos_num}
Table~\ref{tab:pose_arch}(a) studies how the number and temporal placement of key poses affect generation quality.
A consistent trend emerges: fewer anchors yield better FGD across all placement strategies.
With 60-frame clips where the first 8 frames are already fixed as pose prompts, injecting multiple additional anchors over-constrains the model, forcing it to satisfy several hard cues simultaneously and producing unnatural transitions.
Using more anchors consistently degrades FGD, with both 8-anchor and 4-anchor settings performing significantly worse than the single-anchor baseline.
This confirms that drift is a \emph{direction} problem rather than per-frame error accumulation: the model needs one waypoint to aim for, not a dense correction grid.
Figure~\ref{fig:posnum_vis} visualizes this: 8 or 4 anchors introduce visible jitter, while 1 key pose yields smooth motion.

Regarding placement, tail-frame anchors outperform middle and random placements.
Middle-frame anchors may regularize within a clip but cannot prevent drift at clip boundaries, which is where error accumulation occurs in streaming generation.
Tail-frame anchors directly stabilize the transition to the next window, aligning with the closed-loop design of SPG.

\noindent \textbf{Is the Gain from the Database or from Closing the Loop?}
Table~\ref{tab:ablation_combined}(c) dissects anchor quality and refinement strategy.
Replacing retrieved anchors with randomly sampled database poses \emph{substantially worsens} FGD, confirming that arbitrary poses actively hurt generation.
Given retrieved anchors, we compare three strategies: no refinement (directly inserting the anchor), linear refinement (linearly interpolating between the generated and anchor poses for smooth blending), and our flow-matching-based refinement.
Both the no-refinement and linear-refinement baselines underperform our refinement by a clear margin, demonstrating that the gain stems from the combination of nearest-neighbor retrieval and the corrective re-integration from the cached midpoint.
Figure~\ref{fig:fgd_posnum}(b) confirms this visually: no refinement shows pronounced jitter, linear refinement retains discontinuities, whereas our refinement produces a smooth trajectory tracking the ground truth.

\noindent \textbf{Part-aware Architecture.}
Table~\ref{tab:pose_arch}(b) compares single-branch, two-branch, and three-branch DiT variants.
The three-branch design (Hands / Body / Translation) improves FGD over the single-branch baseline with minimal parameter overhead, confirming that separating global displacement from local articulation benefits motion quality.
It is also a structural prerequisite for SPG, which retrieves pose anchors while leaving translation context-dependent.

\subsection{Efficiency Analysis}

\vspace{-4pt}
\begin{center}
    \centering
    \scriptsize
    \setlength{\tabcolsep}{3pt}
    \begin{tabular}{@{}lcccccc@{}}
    \toprule
    \textbf{GPU / DB frames-memory} & \textbf{Init.} & \textbf{FK} & \textbf{Retr.} & \textbf{Ref.} & \textbf{FPS} & \textbf{RT} \\
    \midrule
    V100-16G / 156{,}977-16.6\,MB & 0.460\,s & 0.006\,s & 0.033\,s & 0.230\,s & 76 & yes \\
    5$\times$ / 784k-83\,MB & 0.460\,s & 0.006\,s & 0.132\,s & 0.230\,s & 67 & yes \\
    10$\times$ / 1.57M-166\,MB & 0.460\,s & 0.006\,s & 0.331\,s & 0.230\,s & 54 & yes \\
    20$\times$ / 3.14M-332\,MB & 0.460\,s & 0.006\,s & 0.792\,s & 0.230\,s & 36 & yes \\
    \bottomrule
    \end{tabular}
    \captionof{table}{\textbf{Hardware and database scaling} for one 60-frame, 2s online chunk. StreamTalk remains real time even with a 20$\times$ database.}
    \label{tab:inference_cost}
\end{center}
\vspace{-6pt}

Table~\ref{tab:inference_cost} reports the per-chunk latency on an NVIDIA V100 16GB GPU.
The initial generation and refinement passes are fixed-cost neural inference for a 60-frame chunk, and FK adds only 0.006\,s.
Exact nearest-neighbor retrieval is the only term that scales with the database size.
At the default per-speaker database size, the FK and retrieval overhead is 0.039\,s, while refinement costs roughly half of the initial pass because it starts from the cached midpoint.
When the database is expanded by 20$\times$ to 3.14M frames and 332\,MB, StreamTalk still runs at 36 FPS, which is above the real-time requirement for a 2s chunk.
In practice, a target-speaker database preserves a known speaker style, while a global database provides a fallback when target clips are unavailable.
For larger deployments, indexed nearest-neighbor search can replace exact search without changing the generation pipeline.

\section{Conclusion}

We have presented StreamTalk, a closed-loop framework for streaming co-speech gesture generation.
By formulating clip-wise generation as sequential motion inpainting, StreamTalk introduces a retrieval-based feedback signal---Streaming Pose-Guided Generation (SPG)---that periodically anchors each clip to the plausible motion manifold, transforming the conventional open-loop streaming pipeline into a closed-loop one.
Stochastic Anchor Masking (SAM) bridges the training--inference gap by teaching the model to inpaint complete motion from the same sparse boundary conditions that SPG provides at inference.
A Part-aware DiT further disentangles pose from translation, enabling SPG to retrieve pose anchors without borrowing context-dependent global trajectories.
Experiments on BEAT2 demonstrate state-of-the-art FGD under both single-speaker and multi-speaker settings, with long-horizon stability analysis confirming that SPG effectively suppresses the drift accumulation inherent in open-loop methods.

\section*{Acknowledgments}
This work was supported by the Alibaba Research Intern Program and the Young Scientists Fund of the National Natural Science Foundation of China under Grant No.~624B2110.

%
%
\bibliographystyle{splncs04}
\bibliography{main}
\end{document}